\documentclass{sjtudenglab}

\usepackage{amsmath}
\usepackage{amsfonts}
\usepackage{amssymb}
\usepackage{xcolor}
\usepackage{graphicx}
\usepackage{needspace}

\usepackage{amsmath,amsfonts,bm}

\def\eqref#1{equation~\ref{#1}}

\def\1{\bm{1}}

\DeclareMathAlphabet{\mathsfit}{\encodingdefault}{\sfdefault}{m}{sl}
\SetMathAlphabet{\mathsfit}{bold}{\encodingdefault}{\sfdefault}{bx}{n}

\title{TacForcing: Streaming Action Generation with Execution-Time Tactile Feedback}

\author[1,2,3]{Jianbo Zhou}
\author[4]{Boyuan Zhao}
\author[5]{Yuzheng Zhang}
\author[1,2]{Yiyang Chen}
\author[1]{Wenxin Chen}
\author[6]{Qiuyue Li}
\author[7]{Xiangyang Gu}
\author[8]{Yuhan Cao}
\author[1]{Xiao Xia}
\author[1,2,10]{Yi Yang}
\author[1,9]{Yanzhe Hu}
\author[1,\ddagger]{Zhijie Deng}

\affiliation[1]{School of Computer Science, SJTU} 
\affiliation[2]{Sharpa}
\affiliation[3]{SCUT}
\affiliation[4]{ECUST}
\affiliation[5]{ECNU}
\affiliation[6]{PolyU}
\affiliation[7]{CCUT}
\affiliation[8]{UESTC}
\affiliation[9]{HUST}
\affiliation[10]{SII}

\contribution[\ddagger]{Correspondence to: \url{zhijied@sjtu.edu.cn}}

\abstract{
  Contact-rich manipulation requires adapting to contact states that can evolve substantially within an action horizon. However, chunk-based vision-language-action models predict complete action chunks from observations collected before execution, leaving tactile conditioning stale during execution. Existing tactile-reactive approaches typically rely on separate high-frequency controllers, which increase both architectural and training complexity.
In this paper, we introduce \textbf{TacForcing}, a streaming action-generation framework incorporating execution-time tactile feedback. TacForcing replaces the standard action expert with a streaming expert that generates action blocks sequentially while preserving intermediate states of unfinished blocks. After each block is executed, the expert resumes generation from these states using newly acquired tactile feedback.
To better align tactile conditioning with action execution, we further introduce Execution-Aware Tactile Attention (EATA), which restricts direct access to each tactile update to the next block scheduled for execution. Across six UniVTAC simulation tasks and six real-world contact-rich manipulation tasks on two robot platforms, TacForcing achieves average success rates of 65\% in simulation and 66\% in the real world, outperforming the strongest baselines by 6 and 15 percentage points, respectively.
}

\metadata[Project Website]{\url{https://tacforcing-review.pages.dev/}}
\date{\sffamily\today}

\AtBeginDocument{%
  \fancyheadoffset[R]{-0.3cm}
  \fancyhead[R]{\raisebox{9pt}{\includegraphics[height=35pt]{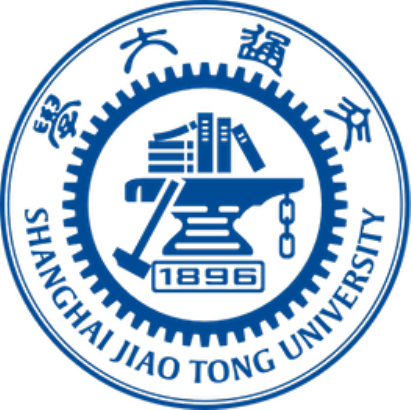}}}
}

\begin{document}

\maketitle

\begin{figure}[!ht]
  \centering
  \includegraphics[width=\linewidth]{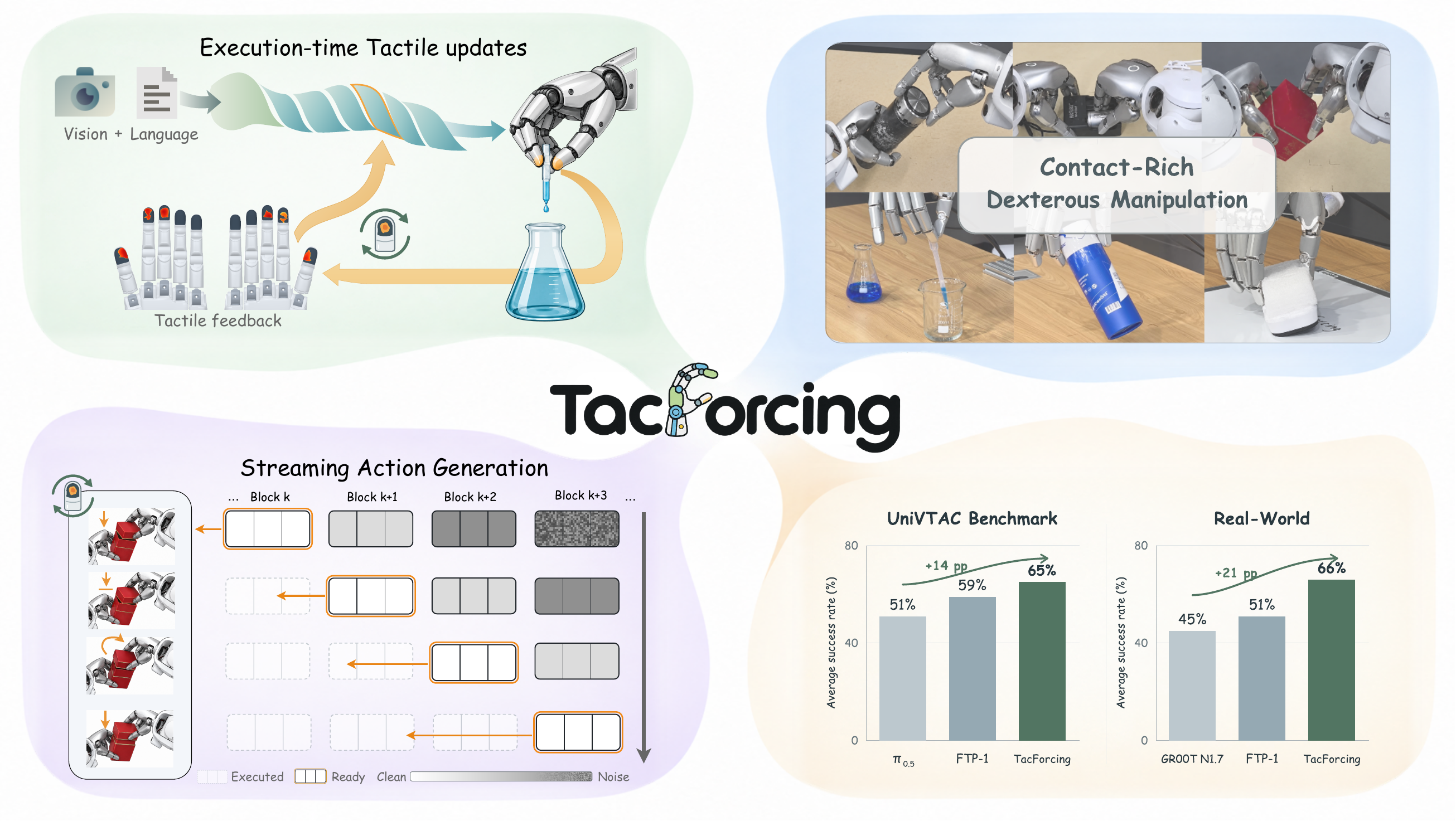}
  \refstepcounter{figure}
  \label{fig:paper_overview}
\end{figure}

\Needspace{8\baselineskip}

\section{Introduction}
\label{sec:introduction}

Contact-rich tasks, including precision assembly, part insertion, and dexterous manipulation, require both semantic understanding and continuous adaptation to evolving contact states \citep{DBLP:conf/corl/HuangWYLL24,DBLP:journals/corr/abs-2607-07287}. Vision-language-action (VLA) models generate robot actions from language instructions and visual observations and have demonstrated broad generalization across manipulation tasks \citep{DBLP:journals/corr/abs-2410-24164,DBLP:journals/corr/abs-2503-14734}. However, visual observations alone cannot reliably reveal changes in force, slip events, or other latent contact states, particularly under occlusion \citep{DBLP:journals/corr/abs-2607-07287}. This perceptual limitation reduces the precision and robustness of contact-rich manipulation.

Recent studies have integrated tactile perception into VLA models for contact-rich manipulation \citep{DBLP:journals/corr/abs-2507-09160,DBLP:journals/ral/ChengZTWZLZS26,DBLP:journals/corr/abs-2606-13102}. 
However, as illustrated in Figure~\ref{fig:visual_tactile_dynamics}, tactile representations can change substantially within a single action chunk even when visual representations remain nearly unchanged. Consequently, conditioning an entire action chunk on a fixed tactile observation creates a growing temporal mismatch between the tactile input and the contact states encountered during execution \citep{DBLP:journals/corr/abs-2503-02881}. 
Existing reactive methods mitigate this mismatch by coupling chunk-level action prediction with dedicated high-frequency tactile pathways \citep{DBLP:journals/corr/abs-2503-02881,DBLP:conf/cvpr/LiCXZFSRD26,DBLP:journals/corr/abs-2606-17055,DBLP:journals/corr/abs-2608-01824}. These pathways enable execution-time adjustments based on newly acquired tactile feedback. However, the additional policy-specific components increase architectural and training complexity, motivating a more integrated approach that incorporates execution-time tactile feedback directly into action generation.

We therefore introduce TacForcing (Figure~\ref{fig:tacforcing_overview}), a streaming action-generation framework that effectively incorporates execution-time tactile feedback. To adapt to evolving contact states, TacForcing partitions each action chunk into sequential blocks and employs a Streaming Action Expert to generate these blocks progressively during execution. Upon completion, each block is dispatched for execution, while the intermediate states of all unfinished blocks are retained for subsequent refinement. Generation then resumes from these states using newly acquired tactile feedback. 
However, streaming generation jointly refines all unfinished blocks, allowing later blocks to directly attend to tactile feedback that may become outdated before their execution. To address this mismatch, we introduce Execution-Aware Tactile Attention (EATA), which restricts direct access to each tactile update to the next block scheduled for execution.

We evaluate TacForcing on six tasks from the UniVTAC simulation benchmark~\citep{DBLP:journals/corr/abs-2602-10093} and six real-world contact-rich manipulation tasks across two robot platforms, comparing against representative vision-only, tactile-conditioned, and tactile-reactive policies. TacForcing achieves average success rates of 65\% in simulation and 66\% in the real world, exceeding the strongest baselines by 6 percentage points in simulation and 15 percentage points in the real world. Ablation studies further show that both execution-time tactile updates and EATA contribute to improved task success. In summary, our contributions are as follows:

\begin{itemize}
    \item We introduce TacForcing, a streaming action-generation framework that effectively incorporates execution-time tactile feedback without a separate reactive controller.

    \item We adapt a Streaming Action Expert to incorporate execution-time tactile feedback and introduce Execution-Aware Tactile Attention (EATA), which reduces temporal mismatch by restricting direct access to each tactile update to the next block to execute.

    \item Experiments demonstrate that TacForcing improves contact-rich manipulation performance across diverse simulated and real-world tasks.
\end{itemize}

\section{Preliminaries and Motivation}
\label{sec:preliminaries}

\subsection{Flow Matching}
\label{sec:flow_matching}

Flow Matching (FM)~\citep{DBLP:journals/corr/abs-2210-02747} learns a conditional velocity field that transports samples from a Gaussian prior to the data distribution. Given condition $y$, we sample $x_1\sim p_{\mathrm{data}}(\cdot\mid y)$, $x_0\sim\mathcal{N}(0,I)$, and $\tau\sim\mathcal{U}(0,1)$, and define the linear interpolation path and target velocity as:
\begin{equation}
    x^\tau=(1-\tau)x_0+\tau x_1,
    \qquad
    u^\star=x_1-x_0.
    \label{eq:fm_path}
\end{equation}
The model is trained by minimizing
\begin{equation}
    \mathcal{L}_{\mathrm{FM}}
    =\mathbb{E}_{y,x_1,x_0,\tau}
    \left[\left\|v_\theta(x^\tau,\tau,y)-u^\star\right\|_2^2\right].
    \label{eq:fm_objective}
\end{equation}
During inference, integrating the learned velocity field $v_\theta$ from $\tau=0$ to $\tau=1$ transforms a sample from the Gaussian prior into a sample from the conditional data distribution.

\subsection{Problem Formulation}
\label{sec:preliminaries_problem_formulation}

We consider language-conditioned, contact-rich robot manipulation. At decision step $t$, the robot receives a visual observation $V_t$, a proprioceptive state $s_t$, a tactile observation $T_t$, and a language instruction $\ell$. Let $c_t=(V_t,s_t,\ell)$ denote the task context. Conditioned on $(c_t,T_t)$, a conventional chunk-based policy models $p_\theta(A_t\mid c_t,T_t)$ and generates an action chunk with horizon $H$,
\begin{equation}
    A_t=(a_t,a_{t+1},\ldots,a_{t+H-1}),
    \qquad
    a_{t+i}\in\mathbb{R}^{d_a},
    \label{eq:action_chunk_preliminaries}
\end{equation}
where $d_a$ denotes the action dimension. In a conventional flow-based action expert, the action chunk $A_t$ and Gaussian noise $\epsilon$ of the same shape serve as the data and prior endpoints, respectively. All action positions share a common flow time and are generated synchronously. The entire chunk is generated from fixed initial inputs $(c_t,T_t)$ before execution begins, so tactile feedback acquired during execution does not update the remaining actions within that chunk.

\subsection{Temporal Mismatch from Fixed Tactile Conditioning}
\label{sec:visual_tactile_dynamics}

\begin{figure}[t]
    \centering
    \includegraphics[width=\linewidth]{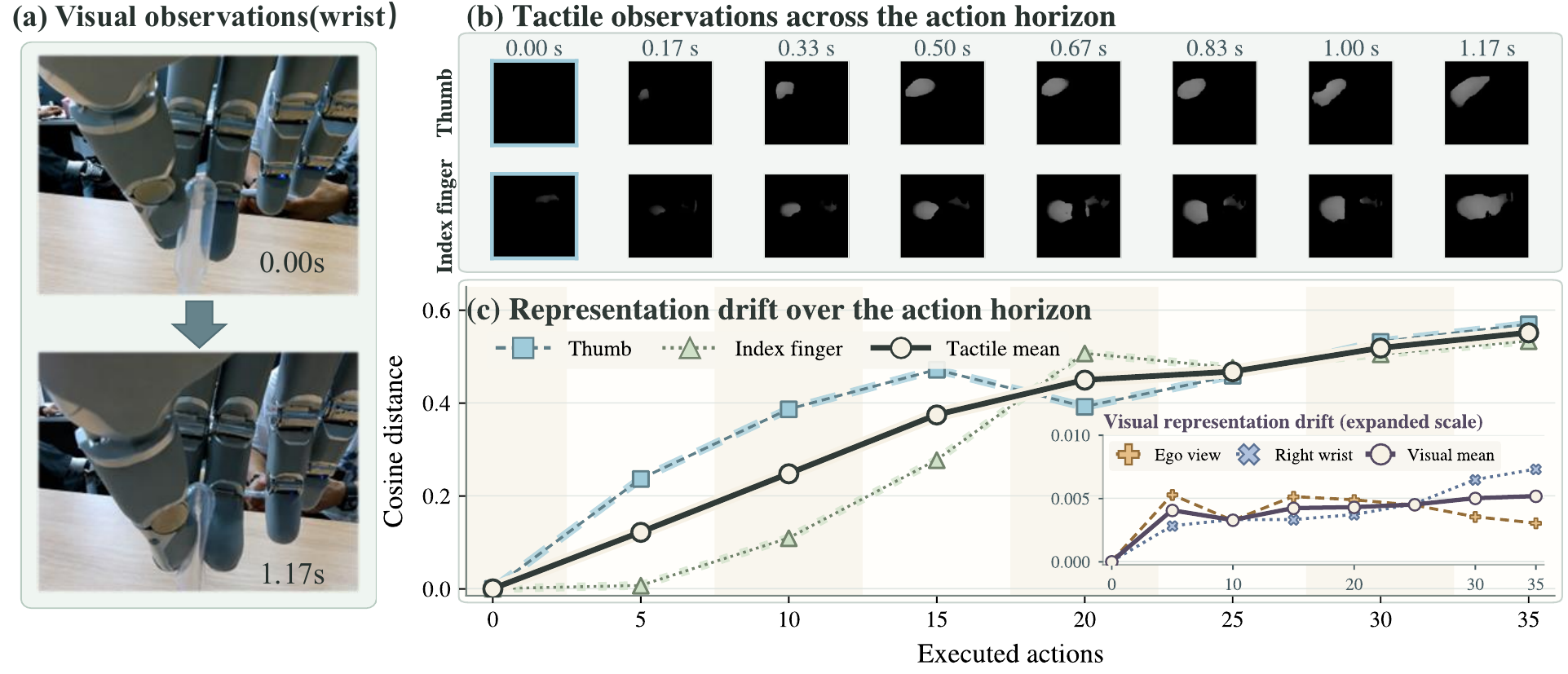}
    \caption{Visual and tactile dynamics during dropper squeezing. (a) Visual observations at the beginning and end of the action horizon. (b) Deformation maps from the thumb and index finger, sampled every five actions. (c) Cosine distances from the initial visual and tactile representations; the inset provides an enlarged view of the visual cosine distances.}
    \label{fig:visual_tactile_dynamics}
\end{figure}

To characterize how visual and tactile observations evolve within an action chunk, we analyze a representative dropper-squeezing episode (Figure~\ref{fig:visual_tactile_dynamics}). Over a 40-action horizon, we sample both modalities every five control steps, yielding eight observations from step 0 to step 35 ($1.17\,\mathrm{s}$ after the initial observation). For each modality, we average distances across its two sensors: the top and wrist cameras for vision, and the thumb and index finger for touch. At step 35, the sensor-averaged cosine distance from the initial representation is approximately $0.005$ for vision and $0.55$ for touch, indicating substantial changes in tactile representations while visual representations remain nearly unchanged in this episode. Details of representation extraction and distance computation are provided in Appendix~\ref{app:representation_dynamics}. This example illustrates that tactile representations can change substantially within an action chunk despite limited changes in visual representations, motivating the use of execution-time tactile feedback rather than relying solely on the initial observation.

\Needspace{9\baselineskip}
\section{Method}
\label{sec:method}

\begin{figure}[t]
    \centering
    \includegraphics[width=\linewidth]{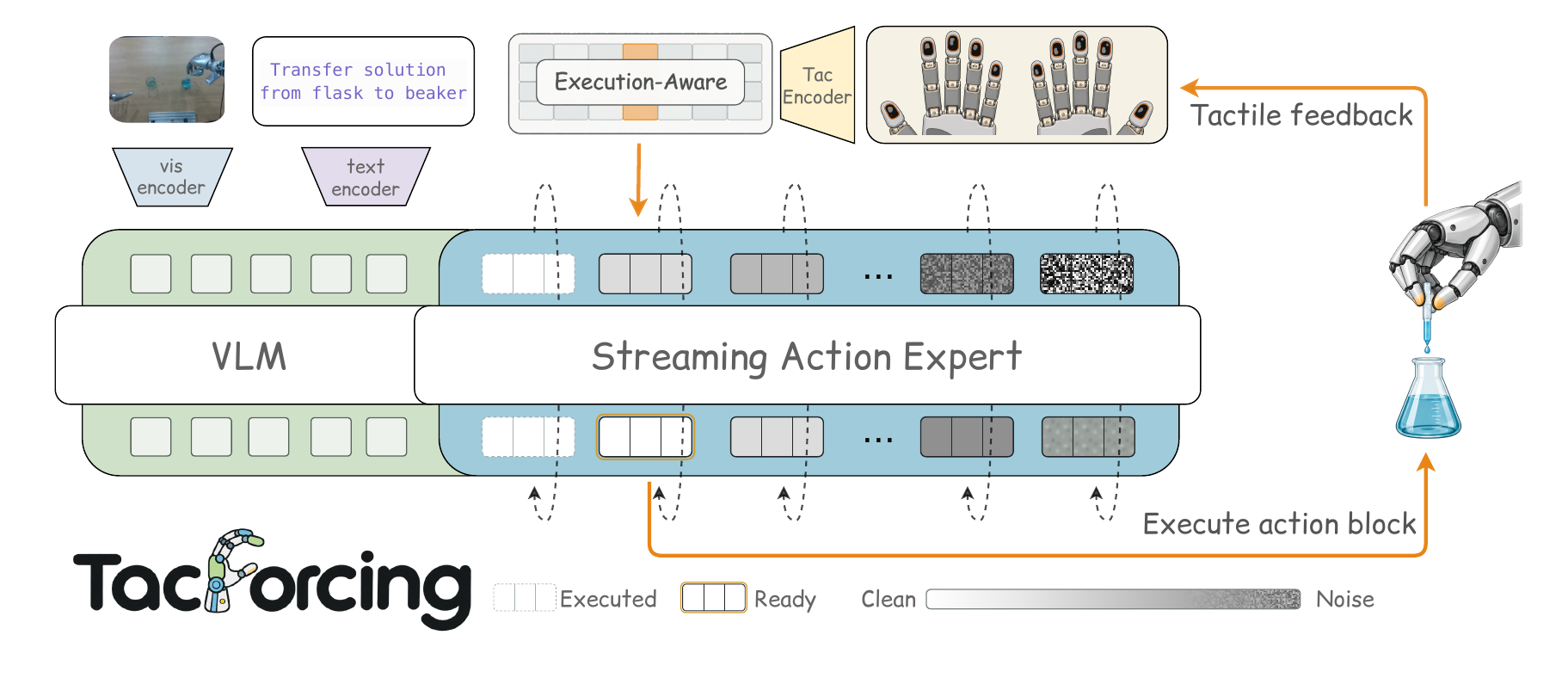}
    \caption{\textbf{Overview of TacForcing}. The VLM encodes the visual observation and language instruction once, and the resulting task context is reused throughout streaming generation. The Streaming Action Expert progressively refines action blocks according to block-specific flow times, allowing successive blocks to become ready for sequential execution. After each ready block is executed, the resulting tactile feedback is encoded before generation resumes from the retained intermediate states of unfinished blocks. Execution-Aware Tactile Attention restricts direct access to the latest tactile feedback to the block scheduled for execution next, thereby reducing the temporal mismatch between tactile acquisition and action execution.}
    \label{fig:tacforcing_overview}
\end{figure}

In this section, we present \textbf{TacForcing} (Figure~\ref{fig:tacforcing_overview}), a streaming action-generation framework that effectively incorporates execution-time tactile feedback. We first introduce block-wise streaming action generation in Section~\ref{sec:streaming_action_generation} and then describe execution-time tactile conditioning in Section~\ref{sec:execution_time_tactile_conditioning}. Finally, we describe the training procedure in Section~\ref{sec:training}.

\subsection{Streaming Action Generation via Block-Wise Flow Scheduling}
\label{sec:streaming_action_generation}

\paragraph{Streaming Action Expert.}
As discussed in Section~\ref{sec:preliminaries_problem_formulation}, a conventional flow-based Action Expert generates the entire action chunk synchronously using only observations acquired before execution. Consequently, tactile feedback obtained during execution cannot condition the actions that remain to be executed. TacForcing addresses this limitation by replacing the standard expert with a Streaming Action Expert that generates the action chunk progressively. Near-term actions are completed and dispatched for execution first, while the intermediate states of later actions are retained for further refinement. This design enables newly acquired tactile feedback to condition the remaining actions within the same chunk without requiring a separate reactive controller.

\paragraph{Block-wise flow scheduling.}
Let $N$ denote the total number of sampling steps, with $n\in\{1,\ldots,N\}$. Building on prior work in streaming generation~\citep{DBLP:conf/nips/0003MDSTS24,DBLP:journals/corr/abs-2603-19199}, we replace the flow time shared across an entire action chunk with position-dependent flow times. At sampling step $n$, we represent these flow times as
\begin{equation}
    \boldsymbol{\tau}^{(n)}
    =\left(\tau_1^{(n)},\tau_2^{(n)},\ldots,\tau_H^{(n)}\right)
    \in[0,1]^H,
    \label{eq:action_flow_times}
\end{equation}
where $\tau_i^{(n)}$ denotes the flow time of the $i$-th action position. We coordinate action completion and tactile updates at the block level, incorporating new tactile feedback after each block is executed.

Assuming $H=KB$ and $N=KS$, we partition the action chunk into $K$ consecutive blocks of $B$ actions and separate successive block completions by $S$ sampling steps. Here, $B$ determines the tactile update interval in executed actions, while $S$ counts numerical integration steps between successive block completions. Let $A_t^{(k)}$ denote the $k$-th block. The function
\begin{equation*}
    b(i)=\left\lfloor\frac{i-1}{B}\right\rfloor+1
\end{equation*}
maps action position $i$ to its corresponding block. All actions in block $k$ share a block flow time $\lambda_k$ and reach completion simultaneously at sampling step $n_k=kS$. The flow-time trajectory is
\begin{equation}
    \lambda_k^{(n)}
    =\min\left(\frac{n}{n_k},1\right),
    \qquad
    k\in\{1,\ldots,K\}.
    \label{eq:block_flow_schedule}
\end{equation}
Accordingly, the flow time of action position $i$ is $\tau_i^{(n)}=\lambda_{b(i)}^{(n)}$. At each sampling step, we use explicit Euler integration to advance every unfinished block, with a block-specific step size $\Delta\lambda_k^{(n)}=\lambda_k^{(n)}-\lambda_k^{(n-1)}$ determined by Equation~\ref{eq:block_flow_schedule}. Completed blocks remain fixed because their step sizes are zero. Since $n_1<n_2<\cdots<n_K$, the blocks become ready for execution sequentially. When $A_t^{(k)}$ completes at $n_k$, it becomes available for execution, while later blocks remain partially generated. Their intermediate states are retained and further refined after $A_t^{(k)}$ is executed.

\subsection{Conditioning on Execution-Time Tactile Feedback}
\label{sec:execution_time_tactile_conditioning}

\paragraph{Block-level tactile updates.}
Within each action chunk, the task context $c_t$ is encoded once as $C_t=f_{\mathrm{ctx}}(c_t)$ and reused throughout chunk generation. In contrast, the tactile condition is refreshed after the execution of each block. Let $T_t^{(k)}$ denote the tactile observation acquired after the first $k$ blocks have been executed, with $T_t^{(0)}=T_t$. The deformation maps from the $M$ fingertips are encoded independently using a shared tactile encoder $f_{\mathrm{tac}}$, producing the tactile tokens
\begin{equation*}
    Z_t^{(k)}
    =\left(z_{t,1}^{(k)},\ldots,z_{t,M}^{(k)}\right).
\end{equation*}
During sampling steps $(k-1)S<n\leq kS$, $Z_t^{(k-1)}$ serves as the latest available tactile representation. After $A_t^{(k)}$ is executed, the newly acquired observation $T_t^{(k)}$ is encoded as $Z_t^{(k)}$ and used to condition the subsequent refinement of the remaining blocks.

\paragraph{Execution-Aware Tactile Attention.}
After the first $k-1$ blocks have been executed, $Z_t^{(k-1)}$ represents the latest contact state and is temporally aligned with $A_t^{(k)}$, the block scheduled to execute next. However, it should not directly condition every unfinished block. Later blocks will be executed only after additional tactile updates and, because contact states can change rapidly as illustrated in Figure~\ref{fig:visual_tactile_dynamics}, may encounter states that differ substantially from the state encoded by $Z_t^{(k-1)}$.

Despite this temporal distinction, the Streaming Action Expert advances all unfinished blocks jointly. Without an additional constraint, unrestricted attention would allow every unfinished action token to attend directly to $Z_t^{(k-1)}$, thereby conditioning later blocks on tactile feedback that may become outdated before their execution. We therefore introduce EATA to restrict direct access to $Z_t^{(k-1)}$ to the action tokens in $A_t^{(k)}$. We implement this restriction by applying an additive mask to the action-to-tactile attention scores, while preserving the expert's remaining attention structure:
\begin{equation}
    \mathcal M_{i,m}^{(k)}=
    \begin{cases}
        0, & b(i)=k,\\[3pt]
        -\infty, & \text{otherwise},
    \end{cases}
    \label{eq:eata_mask}
\end{equation}
where $i\in\{1,\ldots,H\}$ indexes action queries and $m\in\{1,\ldots,M\}$ indexes tactile keys. Under this mask, the current tactile representation directly conditions only the block scheduled to execute next. Later blocks continue to evolve according to the block-wise flow schedule without directly accessing the current tactile tokens and gain direct access to updated tactile feedback when they become the next block to execute. We enforce the same constraint on direct access during training to ensure consistency with the temporal conditioning used during streaming inference.

\subsection{Training}
\label{sec:training}

To match streaming inference, we train the expert on block-wise intermediate states paired with the tactile feedback available at the corresponding execution stage. For each demonstrated action chunk $A_t$, we sample Gaussian noise $\epsilon\sim\mathcal{N}(0,I)$ and a normalized generation progress $p\sim\mathcal{U}([0,1))$. Using $n=pN$ and $N=KS$ in Equation~\ref{eq:block_flow_schedule} yields the flow time and interpolated state of block $k$ as
\begin{equation}
    \lambda_k(p)
    =
    \min\left(\frac{Kp}{k},1\right),
    \qquad
    \widetilde A_t^{(k)}
    =
    \bigl(1-\lambda_k(p)\bigr)\epsilon^{(k)}
    +
    \lambda_k(p)A_t^{(k)}.
    \label{eq:training_block_state}
\end{equation}
The block states form $\widetilde A_t$, with action-wise flow times $\tau_i(p)=\lambda_{b(i)}(p)$ collected in $\boldsymbol{\tau}(p)$.

For the sampled progress $p$, let $k^\star(p)=\lfloor Kp\rfloor+1$. We use the corresponding tactile representation $Z_t^{(k^\star-1)}$ and EATA mask $\mathcal M^{(k^\star)}$. In each demonstration, the selected tactile frame corresponds to offset $(k^\star-1)B$ from the chunk start and precedes execution of block $k^\star$. Let
\begin{equation*}
    \mathcal U(p)
    =
    \left\{
        i\in\{1,\ldots,H\}
        \mid
        \lambda_{b(i)}(p)<1
    \right\}
\end{equation*}
denote the unfinished action positions. For each $i\in\mathcal U(p)$, the predicted velocity is
\begin{equation*}
    \widehat v_{\theta,i}(p)
    =v_{\theta,i}\!\left(
        \widetilde A_t,
        \boldsymbol{\tau}(p),
        C_t,
        Z_t^{(k^\star-1)};
        \mathcal M^{(k^\star)}
    \right).
\end{equation*}
The training objective is
\begin{equation}
    \mathcal L_{\mathrm{train}}
    =\mathbb E_{A_t,\epsilon,p}
    \left[
        \frac{1}{|\mathcal U(p)|}
        \sum_{i\in\mathcal U(p)}
        \left\|
            \widehat v_{\theta,i}(p)
            -\left(a_{t+i-1}-\epsilon_i\right)
        \right\|_2^2
    \right].
    \label{eq:tacforcing_training_objective}
\end{equation}
The loss is applied only to unfinished actions, matching the positions updated at the corresponding stage of streaming inference.

\Needspace{10\baselineskip}
\section{Experiments}
\label{sec:experiment}

We first describe the experimental setup, including the benchmarks, baselines, and implementation details, in Section~\ref{sec:experimental_setup}. We then present the main quantitative results from the simulation benchmark and the real-world platforms in Section~\ref{sec:main_results}. We assess the effects of execution-time tactile conditioning and Execution-Aware Tactile Attention through ablation studies in Section~\ref{sec:ablation_study}.

\subsection{Experimental Setup}
\label{sec:experimental_setup}

\begin{figure}[t]
    \centering
    \includegraphics[width=\linewidth]{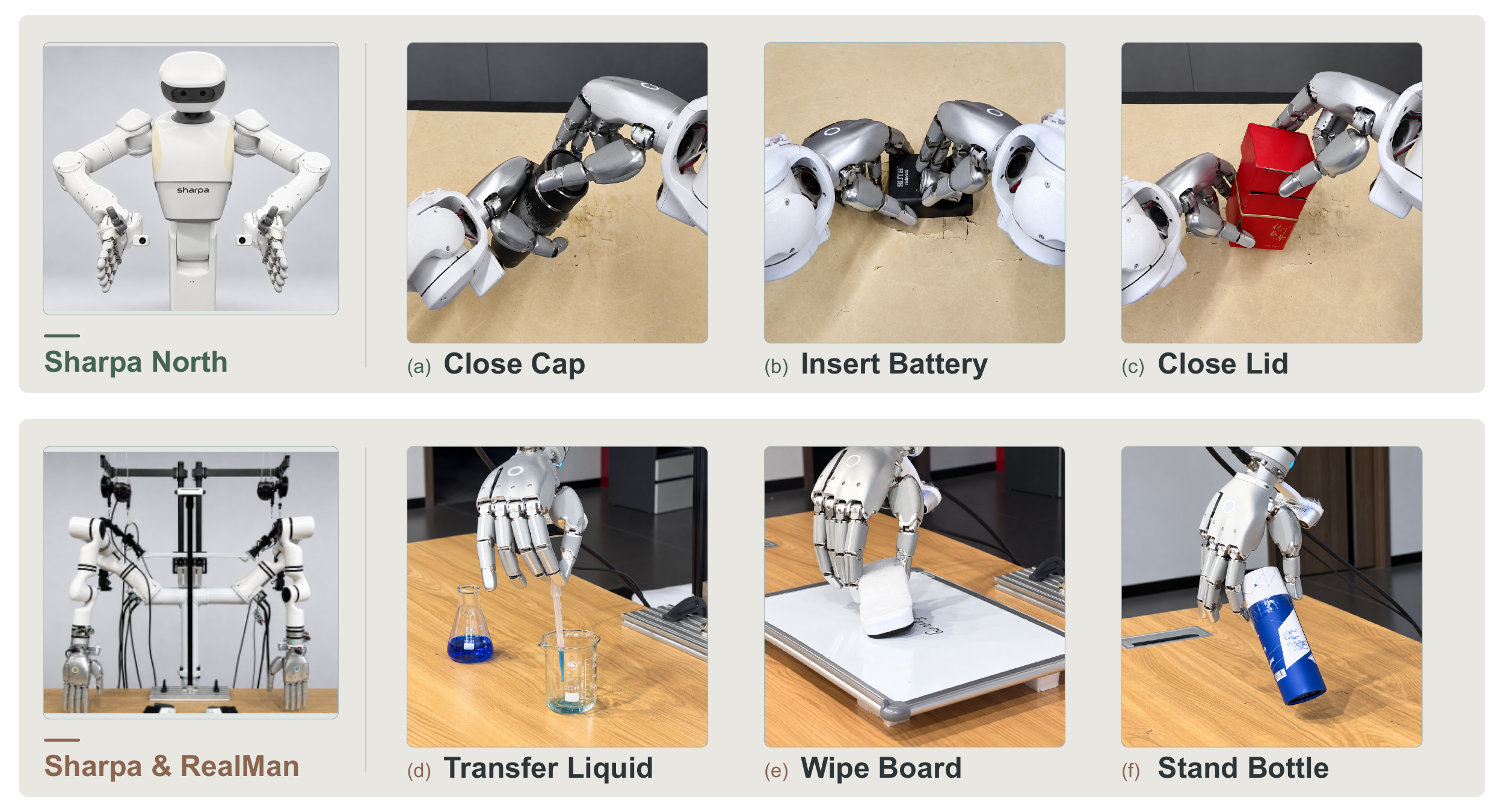}
    \caption{\textbf{Real-world platforms and tasks.} Top: \textit{Close Cap}, \textit{Insert Battery}, and \textit{Close Lid} on Sharpa North. Bottom: \textit{Transfer Liquid}, \textit{Wipe Board}, and \textit{Stand Bottle} on Sharpa \& RealMan. Platform views are shown on the left; task scenes are shown on the right.}
    \label{fig:real_world_main_results}
\end{figure}

\noindent\textbf{Benchmarks.} We evaluate TacForcing on six contact-rich manipulation tasks from the UniVTAC simulation benchmark~\citep{DBLP:journals/corr/abs-2602-10093} and six real-world tasks across two robot platforms. The real-world tasks comprise \textit{Close Cap}, \textit{Insert Battery}, and \textit{Close Lid} on Sharpa North, and \textit{Transfer Liquid}, \textit{Wipe Board}, and \textit{Stand Bottle} on Sharpa \& RealMan (Figure~\ref{fig:real_world_main_results}). Details of the simulation tasks and real-world platforms and tasks are provided in Appendices~\ref{app:simulation_tasks} and~\ref{app:real_world_setup}, respectively.

\noindent\textbf{Baselines.} We compare TacForcing against representative vision-only, tactile-conditioned, and tactile-reactive policies. In simulation, we evaluate four baselines: the vision-only $\pi_{0.5}$~\citep{DBLP:journals/corr/abs-2504-16054}; UniVTAC-ACT~\citep{DBLP:journals/corr/abs-2602-10093}, which augments an Action Chunking Transformer with a pretrained tactile encoder; FTP-1~\citep{DBLP:journals/corr/abs-2606-13102}, which maps heterogeneous tactile observations to unified tokens processed by a shared tactile expert; and RDP~\citep{DBLP:journals/corr/abs-2503-02881}, which combines low-frequency action-chunk prediction with a high-frequency tactile-reactive controller. We use the released checkpoints of UniVTAC-ACT and FTP-1 and train $\pi_{0.5}$ and RDP using their official implementations. In the real world, we evaluate $\pi_{0.5}$ and GR00T N1.7~\citep{DBLP:journals/corr/abs-2503-14734} as vision-only baselines, FTP-1 as a tactile-conditioned baseline, and T-Rex~\citep{DBLP:journals/corr/abs-2606-17055} as a tactile-reactive baseline, all trained using their official implementations. Within each setting, all methods follow the same evaluation protocol.

\Needspace{8\baselineskip}
\noindent\textbf{Implementation Details.} TacForcing is initialized from $\pi_{0.5}$~\citep{DBLP:journals/corr/abs-2504-16054} in simulation and GR00T N1.7~\citep{DBLP:journals/corr/abs-2503-14734} in the real world. In both settings, the tactile encoder is initialized from the pretrained tactile encoder of FTP-1~\citep{DBLP:journals/corr/abs-2606-13102}. We use a block size of $B=5$. In simulation, an action horizon of $H=50$ is divided into $K=10$ blocks. In the real-world setting, we use $H=40$, $K=8$. We report success rates over 100 rollouts per simulation task and 16 independent trials per real-world task. Additional details are provided in Appendix~\ref{app:implementation_details}.

\subsection{Main Results}
\label{sec:main_results}

\begin{center}
    \begin{minipage}{\textwidth}
    \makeatletter
    \def\@captype{table}
    \makeatother
    \centering
    \caption{\textbf{Results on the UniVTAC simulation benchmark.} Success rates (\%). The highest and second-highest distinct values in each column are shown in \textbf{bold} and \underline{underlined}, respectively.}
    \label{tab:simulation_main_results}
    \setlength{\belowcaptionskip}{6pt}
    \setlength{\tabcolsep}{3.5pt}
    \renewcommand{\arraystretch}{1.20}
    \begin{tabular*}{\linewidth}{@{\extracolsep{\fill}}lccccccc@{}}
        \noalign{\hrule height 0.8pt}
        \noalign{\vskip 2pt}
        \textbf{Method} & \shortstack{\textbf{Lift}\\\textbf{Bottle}} & \shortstack{\textbf{Pull-out}\\\textbf{Key}} & \shortstack{\textbf{Lift}\\\textbf{Can}} & \shortstack{\textbf{Put Bottle}\\\textbf{in Shelf}} & \shortstack{\textbf{Insert}\\\textbf{Hole}} & \shortstack{\textbf{Insert}\\\textbf{Tube}} & \textbf{Avg.} \\
        \noalign{\vskip 2pt}
        \noalign{\hrule height 0.5pt}
        \noalign{\vskip 1pt}
        $\pi_{0.5}$ & 88 & \underline{43} & 46 & \textbf{43} & 39 & 48 & 51 \\
        UniVTAC-ACT & 59 & 41 & 24 & 6 & 36 & 58 & 37 \\
        RDP & 84 & 18 & 12 & \underline{41} & 23 & 75 & 42 \\
        FTP-1 & \underline{89} & 35 & \textbf{66} & 23 & \underline{62} & \underline{76} & \underline{59} \\
        \noalign{\vskip 1pt}
        \noalign{\hrule height 0.5pt}
        \noalign{\vskip 1pt}
        \textbf{TacForcing (Ours)} & \textbf{90} & \textbf{48} & \underline{63} & \textbf{43} & \textbf{69} & \textbf{79} & \textbf{65} \\
        \noalign{\vskip 2pt}
        \noalign{\hrule height 0.8pt}
    \end{tabular*}
    \end{minipage}
\end{center}

\phantomsection
\label{sec:simulation_results}
\noindent\textbf{Simulation Results.} As shown in Table~\ref{tab:simulation_main_results}, TacForcing achieves the highest average success rate of 65\%, outperforming FTP-1 at 59\%, the vision-only $\pi_{0.5}$ baseline at 51\%, and the tactile-reactive RDP baseline at 42\%. At the task level, TacForcing ranks first or ties for first on five of the six tasks. On \textit{Insert Hole} and \textit{Insert Tube}, it improves over the strongest baseline, FTP-1, by 7 and 3 percentage points, respectively. On \textit{Lift Can}, it achieves 63\% success, slightly below FTP-1 at 66\%. These results support TacForcing's effectiveness across the evaluated contact-rich tasks.

\begin{table}[!ht]
    \centering
    \setlength{\belowcaptionskip}{6pt}
    \caption{\textbf{Results on six real-world tasks.} All values are success rates (\%). The highest and second-highest distinct rates in each column are shown in \textbf{bold} and \underline{underlined}, respectively.}
    \label{tab:real_world_main_results}
    \setlength{\tabcolsep}{0pt}
    \renewcommand{\arraystretch}{1.15}
    \begin{tabular*}{\linewidth}{@{\extracolsep{\fill}}lccccccc@{}}
        \toprule
        & \multicolumn{3}{c}{\textbf{Sharpa North}} & \multicolumn{3}{c}{\textbf{Sharpa \& RealMan}} & \\
        \cmidrule(lr){2-4}\cmidrule(lr){5-7}
        \textbf{Method} & \makebox[0.09\linewidth]{\renewcommand{\arraystretch}{1.0}\begin{tabular}[c]{@{}c@{}}\textbf{Close}\\[-1pt]\textbf{Cap}\end{tabular}} & \makebox[0.09\linewidth]{\renewcommand{\arraystretch}{1.0}\begin{tabular}[c]{@{}c@{}}\textbf{Insert}\\[-1pt]\textbf{Battery}\end{tabular}} & \makebox[0.09\linewidth]{\renewcommand{\arraystretch}{1.0}\begin{tabular}[c]{@{}c@{}}\textbf{Close}\\[-1pt]\textbf{Lid}\end{tabular}} & \makebox[0.09\linewidth]{\renewcommand{\arraystretch}{1.0}\begin{tabular}[c]{@{}c@{}}\textbf{Transfer}\\[-1pt]\textbf{Liquid}\end{tabular}} & \makebox[0.09\linewidth]{\renewcommand{\arraystretch}{1.0}\begin{tabular}[c]{@{}c@{}}\textbf{Wipe}\\[-1pt]\textbf{Board}\end{tabular}} & \makebox[0.09\linewidth]{\renewcommand{\arraystretch}{1.0}\begin{tabular}[c]{@{}c@{}}\textbf{Stand}\\[-1pt]\textbf{Bottle}\end{tabular}} & \textbf{Avg.} \\
        \midrule
        $\pi_{0.5}$ & 25 & 38 & 13 & 6 & 31 & 44 & 26 \\
        GR00T N1.7 & 44 & \underline{75} & 25 & 19 & 50 & 56 & 45 \\
        FTP-1 & \underline{50} & 63 & 38 & 6 & \textbf{75} & \underline{75} & \underline{51} \\
        T-Rex & 44 & 50 & \underline{44} & \underline{31} & \underline{63} & \underline{75} & \underline{51} \\
        \midrule
        \textbf{TacForcing (Ours)} & \textbf{56} & \textbf{81} & \textbf{50} & \textbf{50} & \textbf{75} & \textbf{81} & \textbf{66} \\
        \bottomrule
    \end{tabular*}
\end{table}

\Needspace{8\baselineskip}
\phantomsection
\label{sec:real_world_results}
\noindent\textbf{Real-World Results.} As shown in Table~\ref{tab:real_world_main_results}, TacForcing achieves the highest average success rate of 66\%, outperforming FTP-1 and T-Rex at 51\% and its vision-only backbone, GR00T N1.7, at 45\%. At the task level, it ranks first on five tasks and ties with FTP-1 on \textit{Wipe Board}. On Sharpa North, TacForcing exceeds the best baseline by 6 percentage points on each of the three tasks. On Sharpa \& RealMan, its largest gain occurs on \textit{Transfer Liquid}, with 50\% success compared with 31\% for T-Rex; it also improves \textit{Stand Bottle} success to 81\%, compared with 75\% for FTP-1 and T-Rex. Overall, these results support the effectiveness of TacForcing across both robot platforms.

\Needspace{10\baselineskip}
\subsection{Ablation Study}
\label{sec:ablation_study}

\begin{figure}[!ht]
    \centering
    \includegraphics[width=\linewidth]{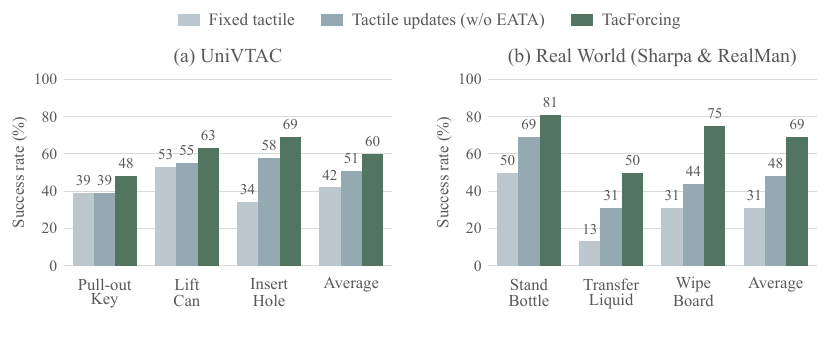}
    \caption{\textbf{Ablation results.} All variants retain streaming generation. Fixed tactile holds the initial tactile observation constant within each action chunk, while tactile updates refresh the observation during execution; both disable EATA. TacForcing enables both updates and EATA. Average denotes the mean success rate over the three tasks in each panel.}
    \label{fig:ablation_results}
\end{figure}

We evaluate execution-time tactile updates and EATA using three variants with the same streaming generation scheme: (i) fixed tactile input without EATA, (ii) tactile updates without EATA, and (iii) full TacForcing. Comparing (i) with (ii) evaluates the benefit of refreshing tactile observations during chunk execution, while comparing (ii) with (iii) assesses the additional contribution of EATA. Figure~\ref{fig:ablation_results} reports results on three simulation tasks and three real-world tasks on Sharpa \& RealMan.

\Needspace{7\baselineskip}
\noindent\textbf{Effect of Execution-Time Tactile Updates.} With EATA disabled, refreshing tactile observations increases the average success rate from 42\% to 51\% in simulation and from 31\% to 48\% in the real world. Compared with fixed tactile input, updates improve \textit{Lift Can} and \textit{Insert Hole} while leaving \textit{Pull-out Key} unchanged, and improve all three real-world tasks. The largest simulation gain occurs on \textit{Insert Hole}, where the success rate increases from 34\% to 58\%. These comparisons support using updated tactile feedback during streaming generation rather than relying on fixed input.

\Needspace{6\baselineskip}
\noindent\textbf{Effect of Execution-Aware Tactile Attention.} With execution-time tactile updates enabled, adding EATA increases the average success rate from 51\% to 60\% in simulation and from 48\% to 69\% in the real world. Success rates increase on all six tasks, with the largest real-world gain on \textit{Wipe Board}, where the success rate rises from 44\% to 75\%. These results highlight the importance of temporal alignment between tactile conditioning and action execution.

\section{Related Work}
\label{sec:related_work}

\paragraph{Diffusion Models for Sequence Generation.}
\label{sec:related_sequence_generation}

Diffusion and flow-based policies generate action chunks through iterative sampling~\citep{DBLP:journals/ijrr/ChiXFCDBTS25,DBLP:journals/corr/abs-2210-02747,DBLP:journals/corr/abs-2410-24164}. Position-dependent denoising supports sequential generation in text, motion, and video~\citep{DBLP:conf/nips/WuFLZGS0LWGDC23,DBLP:conf/siggraph/ZhangLHA24,DBLP:conf/icml/RuheHSH24}. Diffusion Forcing~\citep{DBLP:conf/nips/0003MDSTS24} assigns independent noise levels to sequence tokens, while streaming policies refine future actions as new observations arrive~\citep{DBLP:conf/icra/HoegDE25,DBLP:journals/tmlr/ChenYM025}. FASTER~\citep{DBLP:journals/corr/abs-2603-19199} uses a horizon-aware flow schedule to complete near-term actions earlier and emit them progressively. TacForcing builds on this streaming perspective to incorporate execution-time tactile feedback into retained intermediate action states within the same action chunk.

\paragraph{Tactile-Aware Policies for Contact-Rich Manipulation.}
\label{sec:related_tactile_policies}

Tactile representation learning provides transferable features across sensors and tasks~\citep{DBLP:conf/corl/ZhaoMWA24}. Sparsh~\citep{DBLP:conf/corl/HigueraSBFLKKBL24} learns general-purpose tactile features through self-supervised pretraining, while AnyTouch~\citep{DBLP:conf/iclr/FengHXGSSF025} combines static and dynamic tactile information to support transfer across sensors. Tactile-aware VLAs integrate contact signals into action generation~\citep{DBLP:journals/corr/abs-2507-09160,DBLP:journals/ral/BiMHSS26,DBLP:journals/corr/abs-2606-13102}. Reactive methods combine slower visuomotor planning with faster tactile updates~\citep{DBLP:conf/cvpr/LiCXZFSRD26,DBLP:journals/corr/abs-2606-17055}. RDP~\citep{DBLP:journals/corr/abs-2503-02881} implements this separation through a slow latent diffusion policy and a fast asymmetric tokenizer for tactile feedback control. Predictive and predictive-reactive approaches model future contact states to guide action generation or online refinement~\citep{DBLP:journals/corr/abs-2606-11184,DBLP:journals/corr/abs-2607-07287,DBLP:journals/corr/abs-2608-01824}. TacForcing refines unfinished action blocks using observed tactile feedback, with EATA restricting direct access to each update to the next block scheduled for execution. It requires neither a separate reactive controller nor explicit tactile prediction to adapt actions during chunk execution.
\Needspace{14\baselineskip}
\section{Conclusion}
\label{sec:conclusion}

In this work, we introduced TacForcing, a streaming action-generation framework that effectively incorporates execution-time tactile feedback without a separate reactive controller. TacForcing employs a Streaming Action Expert to generate action blocks progressively during execution, retaining the intermediate states of unfinished blocks and refining them using newly acquired tactile feedback. Execution-Aware Tactile Attention (EATA) restricts direct access to each tactile update to the next block scheduled for execution, thereby reducing the temporal mismatch between tactile acquisition and action execution. Across six simulated UniVTAC tasks and six real-world contact-rich manipulation tasks, TacForcing achieves average success rates of 65\% and 66\%, respectively, outperforming strong baselines in both settings. Ablation results show that both execution-time tactile updates and EATA improve performance, highlighting the value of aligning tactile conditioning with block execution for contact-rich manipulation. The current implementation uses fixed block boundaries and retains visual and proprioceptive inputs within each chunk. Future work could explore adaptive block lengths and the joint scheduling of visual, proprioceptive, and tactile updates.

\clearpage
\setlength{\bibsep}{5pt}
\bibliography{reference}
\bibliographystyle{plainnat}

\newpage
\appendix
\raggedbottom
\section{Additional Implementation Details}
\label{app:implementation_details}

\subsection{Simulation Training Parameters}
We train on 50 demonstration trajectories per task for 15{,}000 steps with a global batch size of 256. We use AdamW with a peak learning rate of $5\times10^{-5}$, a cosine decay to $5\times10^{-6}$, 2{,}000 warm-up steps, a weight decay of $10^{-10}$, and gradient clipping at a global norm of 1.0. We use $H=50$, $K=10$, and $B=5$, with $S=5$ at inference ($N=KS=50$). 

\subsection{Real-World Training Parameters}
We use 100 demonstration trajectories per task, except for \textit{Close Cap} and \textit{Close Lid}, which use 150 and 300 trajectories, respectively. We train for 15 epochs with a global batch size of 256. We use AdamW with a peak learning rate of $6\times10^{-5}$, a cosine decay to $1\times10^{-6}$, 2{,}000 warm-up steps, a weight decay of $10^{-5}$, and gradient clipping at a global norm of 1.0. We use $H=40$, $K=8$, and $B=5$, with $S=2$ at inference ($N=KS=16$).

\subsection{Representation-Dynamics Analysis}
\label{app:representation_dynamics}

We analyze a representative dropper-squeezing episode over a 40-action horizon, sampled every five actions at 30 FPS. Visual representations are extracted separately for the top and wrist cameras by averaging the final spatial tokens from the frozen visual encoder of GR00T N1.7~\citep{DBLP:journals/corr/abs-2503-14734}. Tactile representations are extracted separately for the thumb and index finger from the final output of the tactile encoder in the trained real-world TacForcing model. This encoder is initialized from the pretrained tactile encoder of FTP-1~\citep{DBLP:journals/corr/abs-2606-13102}. Both encoders use their native evaluation preprocessing, with no additional feature normalization before computing cosine distance.

For sensor $j$ and sampling offset $k\in\{0,5,\ldots,35\}$, we compute the cosine distance between the current representation $\mathbf z_k^{(j)}$ and its initial representation $\mathbf z_0^{(j)}$. The modality-level distance is obtained by averaging the resulting distances across the two corresponding sensors:
\begin{equation}
    d_k^{(j)}
    =1-\frac{(\mathbf z_0^{(j)})^\top\mathbf z_k^{(j)}}
    {\lVert\mathbf z_0^{(j)}\rVert_2\lVert\mathbf z_k^{(j)}\rVert_2},
    \qquad
    D_k^{(q)}=\frac{1}{|\mathcal S_q|}\sum_{j\in\mathcal S_q}d_k^{(j)},
    \label{eq:representation_dynamics_distance}
\end{equation}
where $\mathcal S_{\mathrm{vis}}$ contains the top and wrist cameras, and $\mathcal S_{\mathrm{tac}}$ contains the thumb and index finger. The values $0.005$ and $0.55$ reported in the main text are $D_{35}^{(\mathrm{vis})}$ and $D_{35}^{(\mathrm{tac})}$, respectively, at the final sampled offset rather than averages over time.

\Needspace{10\baselineskip}
\section{Real-World Platform and Task Settings}
\label{app:real_world_setup}

\subsection{Real-World Platform}

We evaluate on Sharpa North and Sharpa \& RealMan (Figure~\ref{fig:real_world_main_results}). The Sharpa \& RealMan platform consists of two 7-DoF RealMan RM75 robot arms, each equipped with a 22-DoF Sharpa Wave dexterous hand. Tactile sensors embedded in the fingertips provide deformation maps during contact. A top-mounted RealSense camera captures the global workspace, while wrist-mounted RealSense cameras provide local views of object interactions. Manus Pro data gloves are used in the demonstration-collection interface. Figure~\ref{fig:real_world_platform} shows the complete setup and its main components.

\par\addvspace{10pt}
\noindent\begin{minipage}{\linewidth}
    \makeatletter
    \def\@captype{figure}
    \makeatother
    \centering
    \includegraphics[width=\linewidth]{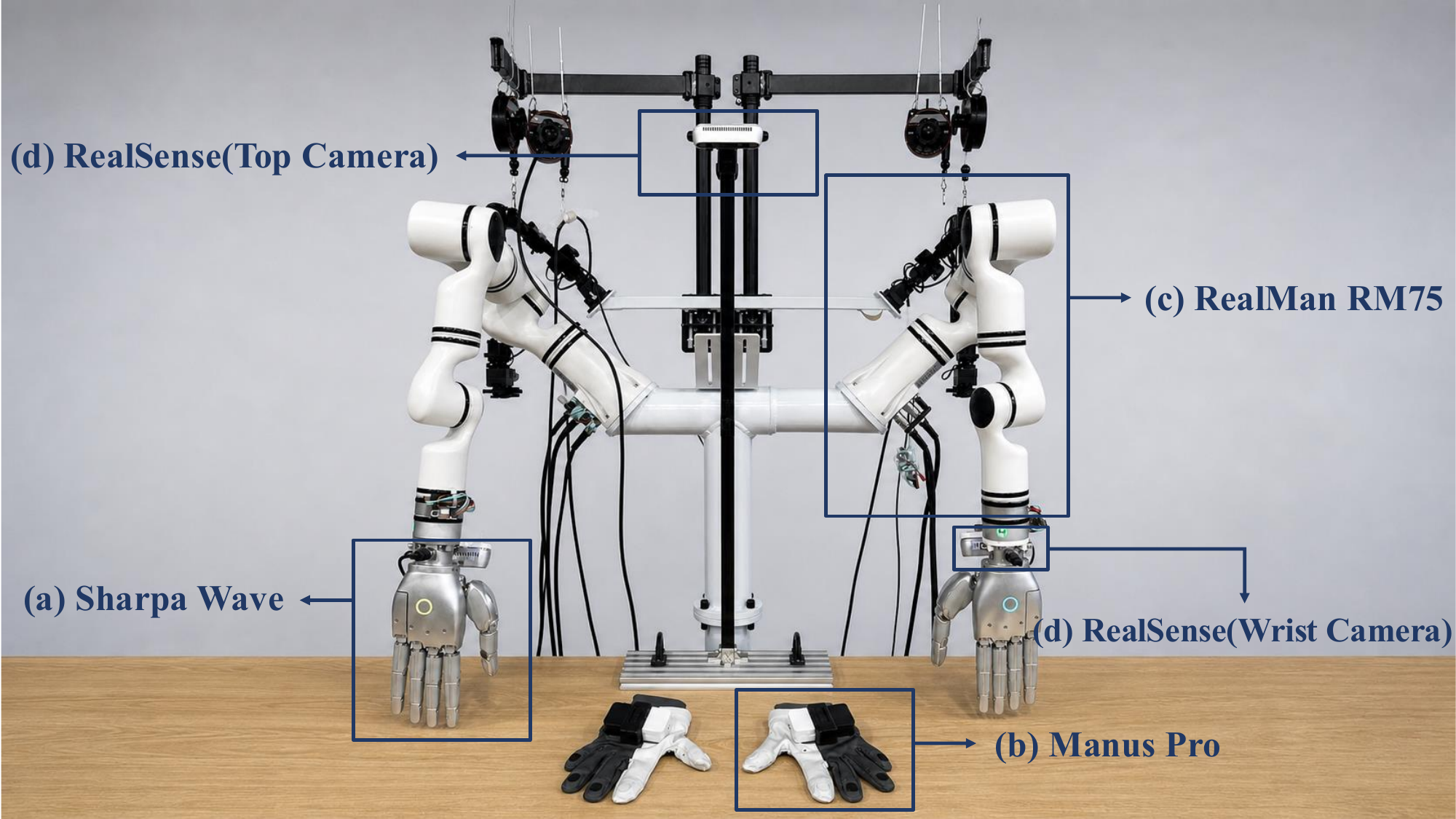}
    \caption{\textbf{Real-world platform.} The setup comprises (a) Sharpa Wave dexterous hands, (b) Manus Pro data gloves, (c) two RealMan RM75 robot arms, and (d) top- and wrist-mounted RealSense cameras.}
    \label{fig:real_world_platform}
\end{minipage}
\par\addvspace{10pt}

\Needspace{8\baselineskip}
\subsection{Real-World Task Settings}

We evaluate six contact-rich manipulation tasks across the two platforms shown in Figure~\ref{fig:real_world_main_results}. \textit{Close Cap}, \textit{Insert Battery}, and \textit{Close Lid} are performed on Sharpa North; \textit{Transfer Liquid}, \textit{Wipe Board}, and \textit{Stand Bottle} are performed on Sharpa \& RealMan. Figures~\ref{fig:sharpa_north_tasks} and~\ref{fig:sharpa_realman_tasks} provide execution sequences for the two platforms in the order listed in Table~\ref{tab:real_world_main_results}. Each method is evaluated over 16 independent trials per task.

\par\addvspace{10pt}
\noindent\begin{minipage}{\linewidth}
    \makeatletter
    \def\@captype{figure}
    \makeatother
    \centering
    \includegraphics[width=\linewidth]{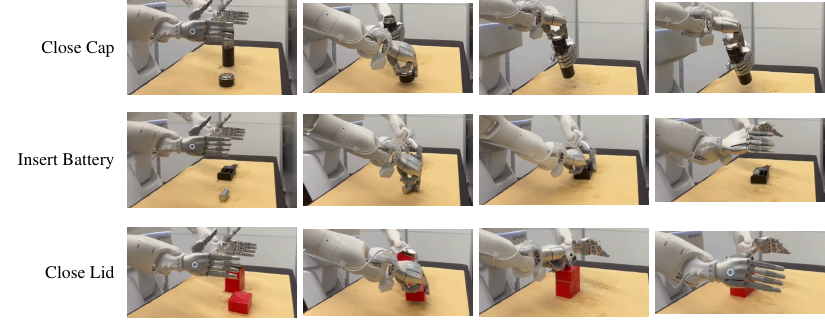}
    \caption{\textbf{Sharpa North tasks.} Representative execution sequences in the task order of Table~\ref{tab:real_world_main_results}. Each row progresses from left to right.}
    \label{fig:sharpa_north_tasks}
\end{minipage}
\par\addvspace{10pt}

\noindent\textit{Close Cap} requires rotating a cap onto a drinking cup; \textit{Insert Battery} requires aligning and seating a battery in its compartment; \textit{Close Lid} requires aligning and pressing the lid onto a tea box.

\par\addvspace{10pt}
\noindent\begin{minipage}{\linewidth}
    \makeatletter
    \def\@captype{figure}
    \makeatother
    \centering
    \includegraphics[width=\linewidth]{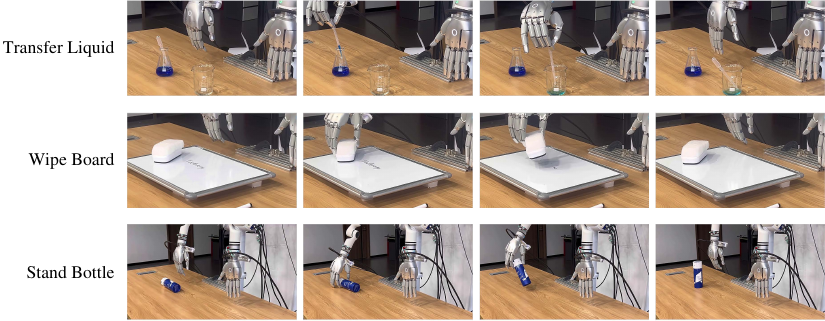}
    \caption{\textbf{Sharpa \& RealMan tasks.} Representative execution sequences in the task order of Table~\ref{tab:real_world_main_results}. Each row progresses from left to right.}
    \label{fig:sharpa_realman_tasks}
\end{minipage}
\par\addvspace{10pt}

\noindent\textit{Transfer Liquid} uses a dropper to move liquid between containers; \textit{Wipe Board} requires sustained contact with a whiteboard; \textit{Stand Bottle} requires reorienting a horizontal bottle and placing it upright.

\Needspace{6\baselineskip}
\section{Simulation Task Details}
\label{app:simulation_tasks}

We evaluate TacForcing on six tasks from the UniVTAC benchmark~\citep{DBLP:journals/corr/abs-2602-10093}. Figure~\ref{fig:simulation_tasks} shows representative execution sequences for these tasks.

\par\addvspace{10pt}
\noindent\begin{minipage}{\linewidth}
    \makeatletter
    \def\@captype{figure}
    \makeatother
    \centering
    \includegraphics[width=0.80\linewidth]{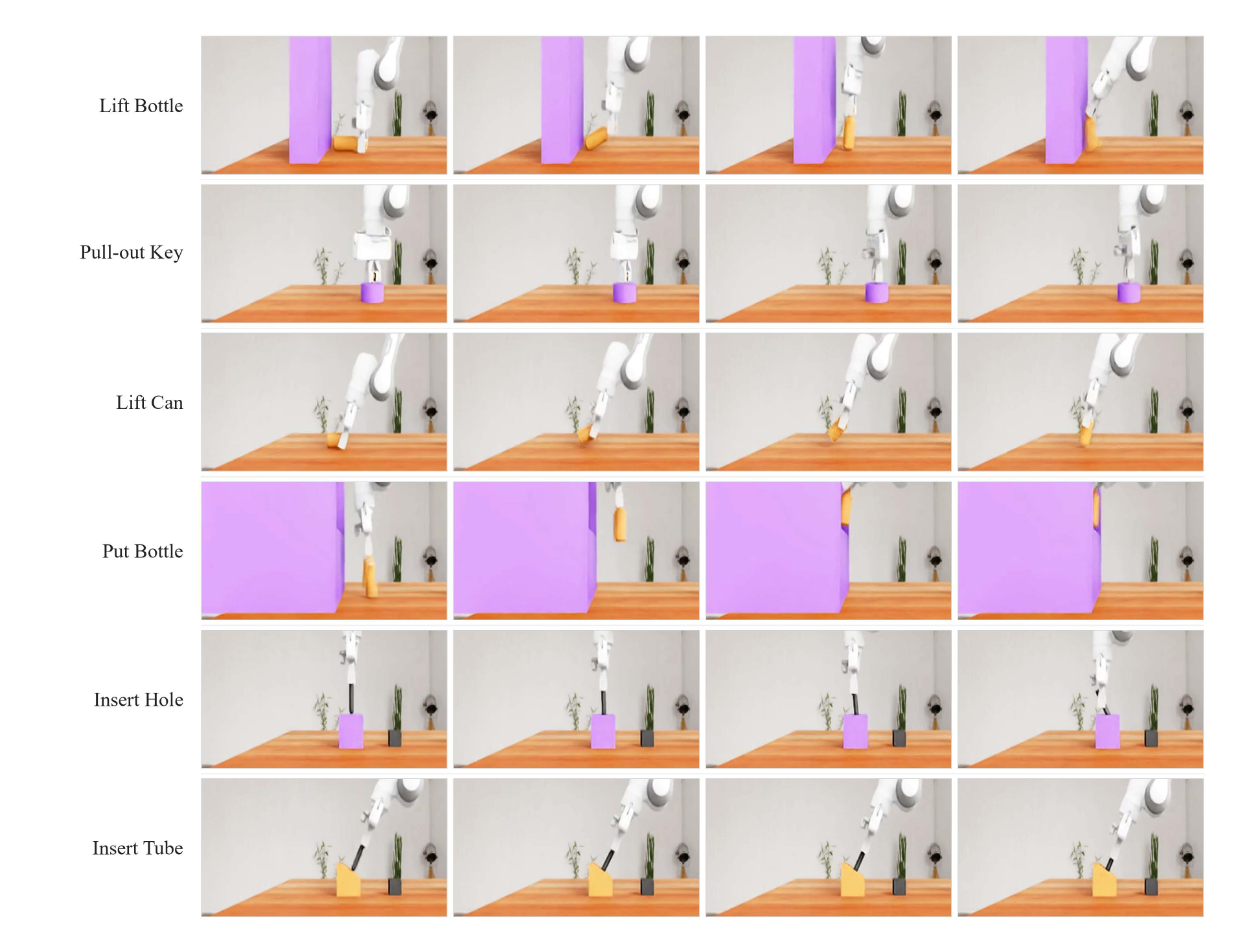}
    \caption{\textbf{UniVTAC simulation tasks.} Representative execution sequences for the six tasks used in our simulation experiments, reproduced from UniVTAC~\citep{DBLP:journals/corr/abs-2602-10093}.}
    \label{fig:simulation_tasks}
\end{minipage}
\par\addvspace{10pt}

\end{document}